\documentclass{article}
\usepackage{ijcai22}

\usepackage{times}
\usepackage{soul}
\usepackage{url}
\usepackage[hidelinks]{hyperref}
\usepackage[utf8]{inputenc}
\usepackage[small]{caption}
\usepackage{graphicx}
\usepackage{amsmath}
\usepackage{amsthm}
\usepackage{booktabs}
\usepackage{algorithm}
\usepackage{algorithmic}
\usepackage{amsfonts,mathtools,amssymb,units,bm,siunitx,textcomp}
\usepackage{graphicx,subfigure}
\usepackage{multirow} 
\usepackage{rotating,makecell,stackengine,xcolor}
\usepackage{tikz}

\usepackage{natbib}

\title{Lifelong Generative Learning via Knowledge Reconstruction}

\author{
	Libo Huang$^1$\and
	Zhulin An$^{1,2}$\footnote{Contact Author}\and
	Xiang Zhi$^{1}$\And
	Yongjun Xu$^{1,2}$
	\affiliations
	$^1$Institute of Computing Technology, Chinese Academy of Sciences\\
	$^2$University of Chinese Academy of Sciences
	\emails
	www.huanglibo@gmail.com,
	anzhulin@ict.ac.cn,
	zhixfor@foxmail.com,
	xyj@ict.ac.cn
}

\begin{document}

\maketitle

\begin{abstract}
Generative models often incur the catastrophic forgetting problem when they are used to sequentially learning multiple tasks, i.e., lifelong generative learning. Although there are some endeavors to tackle this problem, they suffer from high time-consumptions or error accumulation. In this work, we develop an efficient and effective lifelong generative model based on variational autoencoder (VAE). Unlike the generative adversarial network, VAE enjoys high efficiency in the training process, providing natural benefits with few resources. We deduce a lifelong generative model by expending the intrinsic reconstruction character of VAE to the historical knowledge retention. Further, we devise a feedback strategy about the reconstructed data to alleviate the error accumulation. Experiments on the lifelong generating tasks of MNIST, FashionMNIST, and SVHN verified the efficacy of our approach, where the results were comparable to SOTA. 
\end{abstract}

\section{Introduction}
Lifelong learning is an important yet challenging problem, which is notorious for the catastrophic forgetting phenomenon~\citep{kemker2018measuring,parisi2019continual}.
Such a phenomenon reflects the core challenge that the performance dramatically degrades on the tasks learned before, since the model focuses on learning the current task. As the generative replay methods provide an insightful solution in addressing most types of lifelong discriminative learning problems (i.e., task, domain, class lifelong learning) \citep{van2018three,delange2021continual}, how to enable the generative models to lifelong learning has drawn much attention~\citep{lesort2020continual,ramapuram2020lifelong}. This problem is also known as lifelong generative learning.
Some approaches (also known as the pseudo rehearsal) train a new generative model on the mixture data that combines the samples of the current task and the pseudo samples generated from the previous model. These approaches can be fallen into two categories. One is based on the unconditional generative models, while the other is on the conditional generative models.
On the one hand, methods based on the unconditional generative models took the generative adversarial network (GAN)~\citep{shin2017continual} or variational autoencoding (VAE)~\citep{van2018three,van2020brain} to unconditionally accomplish lifelong generative learning. But they are biased towards sampling from the recent task~\citep{seff2017continual,wu2018memory}.
To relieve the biased-sampling problem, on the other hand, conditional generative models were introduced. 
In particular,~\citet{van2020brain} modelled the hidden variable in the VAE as a Gaussian mixture distribution, indicating a mixture of the learned tasks. Then, they balanced the learning of historical tasks and the current one to relieve the biased-sampling problem. \citet{ramapuram2020lifelong} extended the inference capability of VAE from modeling only one hidden variable into two variables, including a continuous feature variable and a discrete task ID variable. Further,~\citet{ye2021lifelong} added one more discrete label ID variable in each task to achieve a better fine-grained category balance. 
For the GAN model, \citet{seff2017continual} only preserved the number of historical labels throughout the lifelong learning process. They creatively embedded the label information of the historical task in the generator model to solve the imbalance problems (i.e., task imbalance and category imbalance). This strategy for GAN is prevalently employed \citep{zhai2019lifelong,liu2020generative}. 

Other approaches train the new generative model on the data from the current task only. 
Referring to \citep{kirkpatrick2017overcoming}, \citet{seff2017continual} enforced the generator to remember the historical knowledge by constraints to update the parameters that are crucial for historical tasks.
Besides,~\citet{wu2018memory} and~\citet{liu2020generative} employed knowledge distillation \citep{hinton2015distilling} to transfer learned knowledge from the previous generator to the new generator.

However, all of the above-mentioned pseudo rehearsal methods train the generative model partly on the pseudo data, which own the uncertain quality from the sampling, and thus result in the~\textit{error accumulation} problem. 
In addition, methods trained only on the current data are merely investigated in the GAN framework, and the~\textit{time consumption} of generating suitable instances is extensive. Compared with GAN, VAE not only provides a stable training mechanism but keeps satisfactory sample diversity~\citep{ramapuram2020lifelong}.

In this paper, motivated by the conditional GAN (CGAN), we first develop a variant of the conditional VAE (CVAE), which enjoys high efficiency in the training process. 
To ensure remembering the previous knowledge, inspired by the intrinsic reconstruction character in the VAE, we particularly propose a knowledge reconstruction loss to guide the decoder training. 
Further, we devise a feedback strategy about the reconstructed data to encourage CVAE to encode the reconstructed sample consistent with the real one. 
Finally, we present a lifelong generative learning algorithm via knowledge reconstruction (LGLvKR in short), which is only trained on the current data without error accumulation from the pseudo sampling. 
Compared with CGAN-based methods, our algorithm LGLvKG obtains comparable results with less computational resources.

\section{Background and Problem Definition}
In this section, we first introduce the concepts used throughout this paper. They include the generalized conditional generative model (GCGM), and a variant of CVAE. We then formalize the definition of our lifelong generative problem.

\subsection{Genelized Conditional Generative Model} \label{subsec:gcgm}
Consider an observed dataset with $N$ conditional samples $\mathcal{D}=\{(x_i, y_i)|i=1,..., N\}$ where $x_i$ and $y_i$ indicate the image and the label for the $i^{th}$ sample, respectively. GCGM aims to estimate the distribution of each label in $y$ based on observations. With estimated distributions, real data with the specified label are well sampled, i.e., conditional sample generation.

\begin{figure}[t]
	\centering
	\subfigure[GCGM]{\label{fig:fig1a}\includegraphics[width=0.43\linewidth]{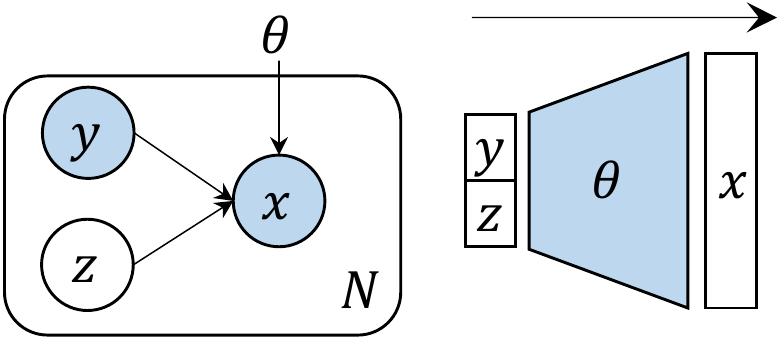}}
	\subfigure[CVAE]{\label{fig:fig1b}\includegraphics[width=0.55\linewidth]{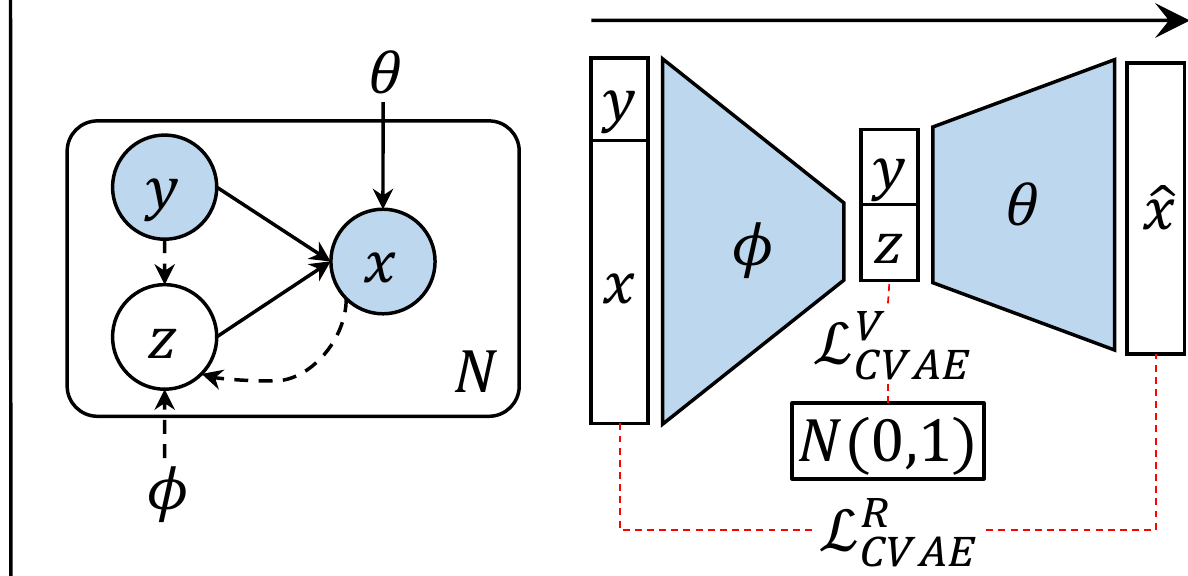}}
	\caption{Schematics of the conditional generative networks, i.e., the graphical model (left) and network architecture (right). (a) GCGM shows how the observed variable, $x$, is generated by an unobserved latent variable $z$ and another observed variable $y$. GCGM aims to get the parameter $\theta$ so that one can sample $x$ with specific characteristics as indicated in $y$. (b) In CVAE, first $x$ and $y$ are encoded to a latent variable $z$ which is expected to be standard Gaussian. Next, $y$ is again used with $z$ to the decoder to reconstruct a variable $\hat{x}$. We expect variables $x$ and $\hat{x}$ are as same as possible.}
	\label{fig:fig1}
\end{figure}

Concretely, as the graphical model shown in Fig.\ref{fig:fig1a}, GCGM expects to estimate a conditional probability distribution with the parameter, $\theta$, through the observable variable $y$ and the unobservable variable $z$. 
In the network architecture,
the generative model with the parameter $\theta$ enables us to generate $x$ conditioned by $y$ with a prior $z$. 
To get a proper $\theta$, the maximum logarithmic likelihood estimation is typically introduced:
\begin{align} \label{eq:1}
\max_{\theta }\log p_{\theta }( x|y) =\max_{\theta }\log\int p_{\theta }( x|y,z) p( z)\mathrm{d} z.
\end{align}
To make Eq.(\ref{eq:1}) integrally tractable~\citep{kingma2014auto}, we suggest a novel conditional variational autoencoder shown in Section~\ref{subsec:2.2}.

\subsection{Conditional Variational Autoencoder} \label{subsec:2.2}
The existing CVAE architecture embeds the conditional information, $y$, either in the encoder~\citep{sohn2015learning} or in the decoder~\citep{kingma2014semi}. Unlike them, we use a variant of CVAE where $y$ is embedded in both encoder and decoder, parametrized by $\phi$ and $\theta$, respectively, as shown in Fig.\ref{fig:fig1b}. 
In this way, we utilize the conditional information further. Using the Jensen's inequality~\citep{ramapuram2020lifelong}, the evidence lower bound (ELBO) is,
\begin{align} \label{eq:2}
&\log p_{\theta }( x|y) =\log\int \frac{q_{\phi }( z|y,x)}{q_{\phi }( z|y,x)} p_{\theta }( x|y,z) p( z)\mathrm{d} z, \nonumber \\
&\geq \mathbb{E}_{q_{\phi}(z|y,x)}[\mathrm{\log} p_{\theta}(x|y,z)] -\mathrm{KL}\left[q_{\phi}(z|y,x) \| p(z)\right],
\end{align}
where $\mathbb{E}_{q_{\phi}(z|y,x)}[\mathrm{\log} p_{\theta}(x|y,z)]$ means to take the expectation of $\mathrm{\log} p_{\theta}(x|y,z)$ with respect to the distribution $q_{\phi}(z|y,x)$, $\mathrm{KL}\left[q_{\phi}(z|y,x) \| p(z)\right]$ is the Kullback–Leibler divergence between the posteriori distribution $q_{\phi}(z|y,x)$ and the prior distribution $p(z)$ \citep{bishop2006PRML}. In this paper, we consider the prior $p(z)$ as standard Gaussian distribution $N(0, 1)$. For more priors, please refer to~\citep{kingma2014semi,kingma2014auto}. 

Therefore, a proper conditional data generation model $p_{\theta}(x|y,z)$ and a conditional encoding model $q_{\phi}(z|y,x)$ can be obtained by back-propagating the following CVAE loss based on the observed data set $\mathcal{D}$,
\begin{align} \label{eq:3}
& \mathcal{L}_{CVAE}(\phi,\theta) = \mathcal{L}_{CVAE}^{R}(\phi,\theta) + \mathcal{L}_{CVAE}^{V}(\phi), \\
&= -\mathbb{E}_{q_{\phi}(z|y,x)}[\mathrm{\log} p_{\theta}(x|y,z)] + \mathrm{KL}\left[q_{\phi}(z|y,x) \| p(z)\right]. \nonumber
\end{align}
Here, for convenience, we named $\mathcal{L}_{CVAE}^{R}$ and $\mathcal{L}_{CVAE}^{V}$ as the reconstruction error and variational error, respectively. By minimizing $\mathcal{L}_{CVAE}(\phi,\theta)$, the conditional encoder $q_\phi(z|y,x)$ promotes $x$ and $y$ to be the standard Gaussian distributions $N(0, 1)$ while the conditional decoder $p_\theta(x|y,z)$ reconstructs $y$ and $z$ back to the original sample $x$.

\subsection{Lifelong Generative Problem Definition} \label{sebsec:2_3}
Lifelong generative learning extends the single-distribution estimation task mentioned in Section~\ref{subsec:gcgm} to a sequence of $T$ tasks\footnote{For simplicity, we use $p_{\theta_t}(x^t|y^t)$ to replace 
$p_\theta(x|y)$ in Eq.\eqref{eq:1} if handling the $t^{th}$ task with the $t^{th}$ distribution, $t=1,...,T$.}.
Concretely, each task characterized by a distinct observed dataset $\mathcal{D}^t=\{(x_i^t, y_i^t)|i=1,...,N_t\}$, $t=1,...,T$, sampled from the desired distributions, $p_{\theta_t}=p_{\theta_t}(x^t|y^t)$. 
Lifelong generative models aim to estimate the true distribution $p_\theta=\int\prod_{t=1}^T p_{\theta_t}\left(x^t|y^t ,z\right) p(z)\mathrm{d} z$ related the whole tasks learned so far although the tasks occur in an sequential manner.
\begin{figure}[!h]
	\centering
	\includegraphics[width=0.37\textwidth]{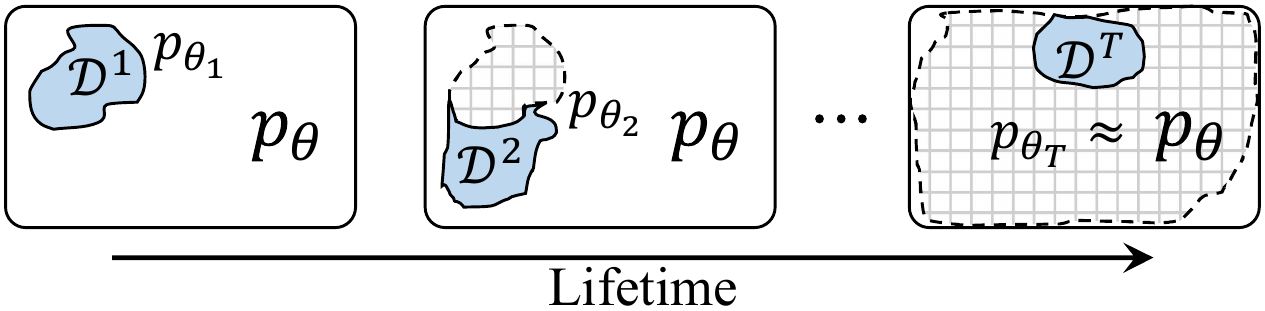}
	\caption{
	A demo of lifelong generative problems. Here, there is only one accessible dataset, $\mathcal{D}^i, i=1,...,T$, indicated by the shaded area for each current task. The grid-filled area represents the distribution that has been learned before. Eventually, the true distribution, $p_\theta$, is accumulated learned after the lifetime of learning.}
	\label{fig:fig2}
\end{figure}

As shown in Fig.\ref{fig:fig2}, during learning the second task with the only accessible dataset $\mathcal{D}^2$, the model not only needs to well estimate the distribution of $\mathcal{D}^2$ but also remember the learned distribution of $\mathcal{D}^1$. Following this lifetime accumulated learning, the final estimated distribution $p_{\theta_T}$ with only accessible dataset $\mathcal{D}^T$ is required to be the same as the true distribution $p_\theta$.

\section{Proposed LGLvKR}
In this section, we propose LGLvKR to well solve the lifelong generative problem mentioned in Section~\ref{sebsec:2_3}. 
Concretely, we first extend the intrinsic reconstruction character of VAE to the retention of historical knowledge. Then, the feedback consolidation strategy is conducted on the reconstructed data to ensure the superior performance. Finally, we give the whole training process and its pseudo code about our proposed LGLvKR.
\subsection{Knowledge Reconstruction} \label{subsec:knowledge}
\begin{figure}[t]
	\centering
	\subfigure[Graphical model]{\label{fig:fig3a}\includegraphics[width=0.42\linewidth]{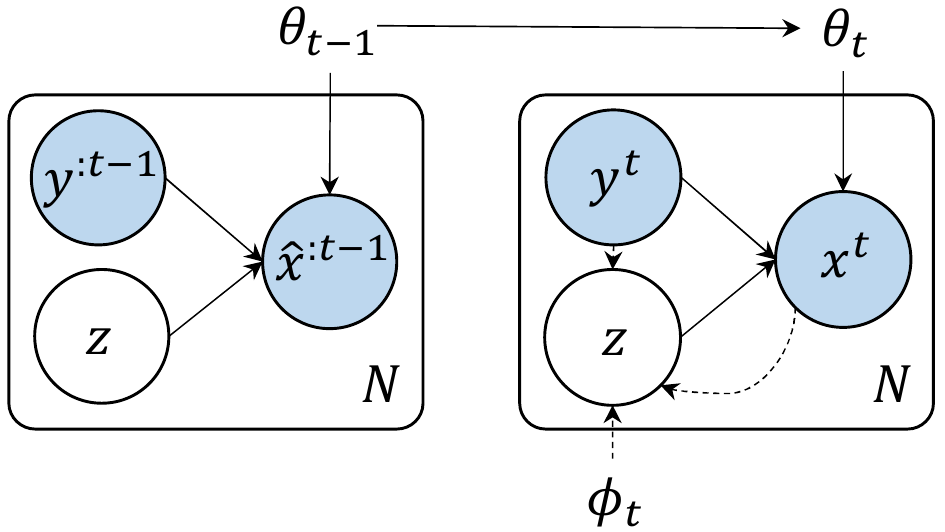}}
	\subfigure[Network architecture]{\label{fig:fig3b}\includegraphics[width=0.57\linewidth]{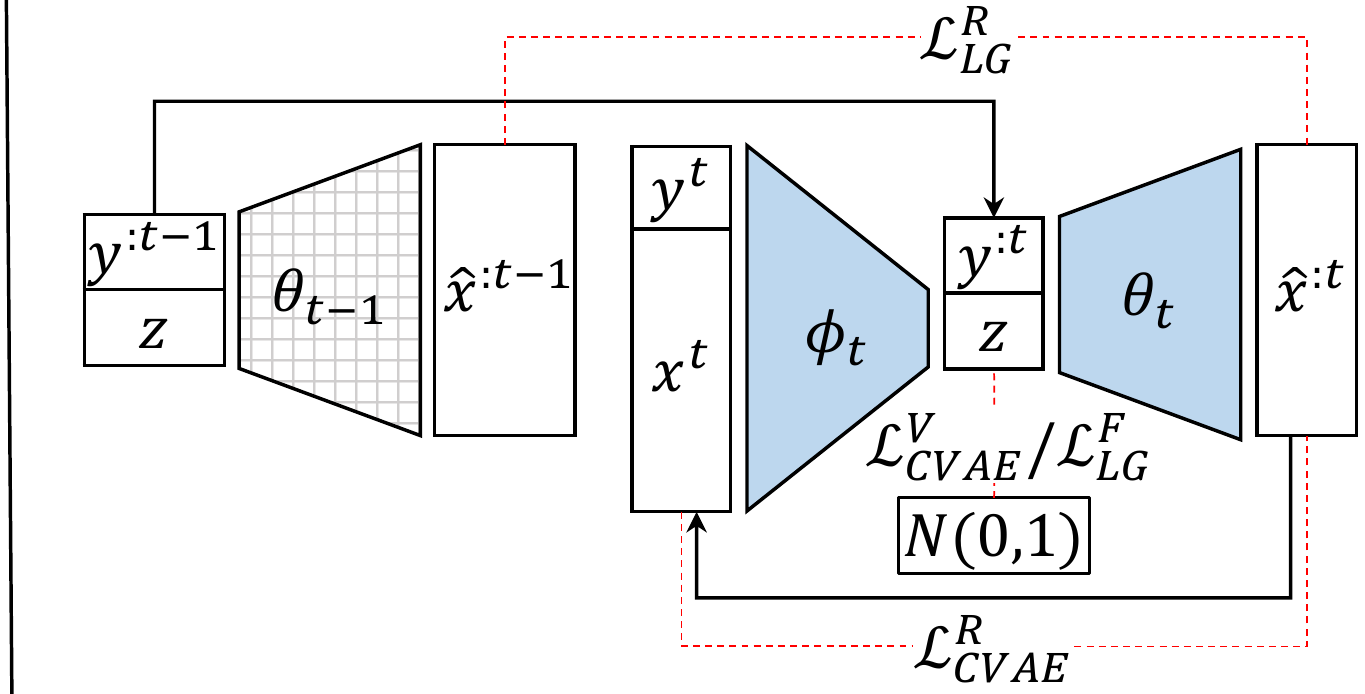}}
	\caption{Schematics of the proposed LGLvKR network.}
	\label{fig:fig3}
\end{figure}

Recall from Section~\ref{sebsec:2_3} that the key to solving the lifelong generative problem is how to retain the historically learned distributions when estimating the current task.
As shown in the left-hand side of Fig.\ref{fig:fig3a}, we claimed that the learned distribution is well retained in the parameter $\theta_{t-1}$. Here, $y^{:t-1}$ in Fig.\ref{fig:fig3} indicates the labels of tasks learned so far.

Instead of training on the mixture data that are partly from the current task and partly generated from the previous model to retain the historical knowledge~\citep{ramapuram2020lifelong,ye2021lifelong},  we extend the intrinsic reconstruction character of VAE to the knowledge reconstruction.
Specifically, as shown in Fig.\ref{fig:fig3b}, the historical decoder $p_{\theta_{t-1}}$ is frozen and saved after training on the dataset $\mathcal{D}^{t-1}$, and the knowledge about the learned tasks (i.e., from task $1$ to $t-1$) is retained in $\theta_{t-1}$. Note that the learned labels are accumulated and also saved.
By inputting the current decoder, $p_{\theta_t}$, and the historical decoder, $p_{\theta_{t-1}}$, with the same latent variable $z$ and labels $y^{:t-1}$, we restrict their reconstruction output to be as consistent as possible. In this way, $p_{\theta_t}$ could well reconstruct the historical knowledge retained in the $p_{\theta_{t-1}}$. 
Therefore, we define a knowledge reconstruction loss associated with the lifelong generation as: 
\begin{align}
&\mathcal{L}_{LG}^{R}(\theta_t)  \\
&= -\mathbb{E}_{p_{\theta_{t-1}}(\hat{x}|y,z),y\sim U\left(1_y, (t-1)_y\right),z\sim N(0,1)}\left[\log p_{\theta_t}(\hat{x}|y,z)\right], \nonumber
\end{align}
where $U(a, b)$ is the discrete uniform distribution, $1_y$ and $(t-1)_y$ indicate the unique labels contained in $y^1$ and $y^{t-1}$, respectively\footnote{In this paper, we assumed the labels related to the whole $T$ tasks are well organized and disjointed, which is common in lifelong learning~\citep{belouadah2021comprehensive}. Therefore, $y\sim U\left(1_y, (t-1)_y\right)$ means uniformly sampling $y$ from the labels learned so far.}. 
It is worth noting that such a knowledge reconstruction loss about each sample has two inputs: a latent $z$ sampled from the prior distribution, and a label-attribute $y$ sampled from the distribution $U\left(1_y, (t-1)_y\right)$.
Given $z$ and $y$, we can generate a sample from the frozen historical decoder $p_{\theta_{t-1}}(\hat{x}|y,z)$, and subsequently compare this sample with the one generated by the trainable $p_{\theta_{t}}(\hat{x}|y,z)$ with the same inputs.  
We take consistently the same comparing strategy as the one used in CVAE reconstruction loss.

\subsection{Feedback Consolidation}

We assume in Section~\ref{subsec:knowledge} that the froze historical decoder $p_{\theta_{t-1}}$ has optimal parameters. Then, the main reason for $p_{\theta_{t}}$ not remembering the considerable knowledge is due to the poor generation by the decoder $p_{\theta_{t-1}}$. 

To guarantee the generation performance of the decoder, inspired by~\citet{verma2018generalized}, we introduce a feedback consolidation strategy.  
Specifically, as shown in Figure \ref{fig:fig3b}, when training CVAE, the encoder is used twice to ensure that the distribution obtained from the current generated data $\hat{x^t}$ follows the true sampling distribution.
Therefore, we formally design a feedback consolidation loss associated with the lifelong generation as:
\begin{align} \label{eq:5}
\resizebox{0.91\linewidth}{!}{$
\mathcal{L}_{LG}^{F}(\phi_t, \theta_t) =-\mathbb{E}_{p_{\theta_t}\left(\hat{x}^{t} |y^{t} ,z\right)}\left[ \mathrm{KL}\left[ q_{\phi_t}\left(z|y^t ,\hat{x}^t\right)\bigl{\|} p(z)\right]\right]$.}
\end{align}
Note that the feedback consolidation is only devoted to the reconstructed data related to the current task $t$. 

Hence, accompanying the CVAE loss with the knowledge reconstruction loss as well as the feedback consolidation loss, the overall learning objective is defined as:
\begin{align} \label{eq:6}
\min_{\phi_t, \theta_t}\left\{ \mathcal{L}_{CVAE}\left(\phi_t,\theta_t\right) + \lambda_t^r\cdot\mathcal{L}_{LG}^R\left(\theta_t\right) + \lambda^f_t\cdot\mathcal{L}_{LG}^F\left(\phi_t, \theta_t\right) \right\},
\end{align}
where the weights, $\lambda^r_t>0$ and $\lambda^f_t>0$, are hyperparameters.

\subsection{Training LGLvKR}
In this section, we summarize the whole training process of LGLvKR when encountering multiple sequential tasks. Look back to Fig.\ref{fig:fig2}, on the one hand, we could only access the observed dataset $\mathcal{D}^1=\{\left(x_i^1, y_i^1\right)|i=1,...,N_1\}$ when encountering the first task. At this stage, we train the proposed CVAE model with the below objective function augmented only with the feedback consolidation loss,
\begin{align} \label{eq:7}
\min_{\phi_1, \theta_1}\left\{ \mathcal{L}_{CVAE}\left(\phi_1,\theta_1\right) + \lambda_1^f\cdot\mathcal{L}_{LG}^F\left(\phi_1, \theta_1\right) \right\},
\end{align}
where $\lambda_1^f=1$. After the training, the decoder model $p_{\theta_1}$ is frozen and saved with the unique labels contained in $y^1$.

On the other hand, when encountering the next task characterized by the observed dataset $\mathcal{D}^2=\{\left(x_i^2, y_i^2\right)|i=1,...,N_2\}$, a snapshot about $p_{\theta_1}$ is taken to initialize $p_{\theta_2}$ before learning. Note that, the historical encoder $p_{\theta_1}$ is frozen while the current encoder $p_{\theta_1}$ is trainable throughout the training stage. At this stage, we train the proposed CVAE model using Eq.\eqref{eq:6} with $t=2$. Since we have saved the unique labels in $y^1$, the knowledge reconstruction loss, $\mathcal{L}_{LG}^{R}$, could be well calculated. After learning this task, again, the current decoder model $p_{\theta_2}$ is frozen and saved along with the accumulated unique labels contained in $y^{:2}$ (i.e., $y^1 \cup y^2$). 
Along with this iterative way, we train LGLvKR until finishing learning the whole sequential $T$ tasks. 
Here, the weights $\lambda_t^f=1, t=1,...,T$ while $\lambda_t^r=t-1, t=1,...,T$. 
We summarize our method LGLvKG in Algorithm~\ref{alg:1}.
\begin{algorithm}[tb]
	\caption{LGLvKR}
	\label{alg:1}
    \textbf{Input}: A sequence of $T$ datasets $\mathcal{D}^t,\ t=1,...,T$.\\
	\textbf{Parameter}: $\{\lambda_t^f\ |\ t=1,...,T\}$. \\
	\textbf{Output}: Decoder, $p_{\theta_T}$, and the set of unique labels, $\{y^{:T}\}$.
	
	\begin{algorithmic}[1] 
		\STATE Observe the dataset $\mathcal{D}_1$.
		\STATE $\{\phi_1, \theta_1\}\leftarrow$ Update CVAE using Eq.\eqref{eq:7} with $\mathcal{D}_1$.
		\STATE Save the decoder, $p_{\theta_1}$, and the labels, $\{y^1\}$.
		\FOR{$t = 2,...,T$}
		\STATE $\{\theta_t\}\leftarrow$ Initialize the current decoder with $\{\theta_{t-1}\}$.
		\STATE Observe dataset $\mathcal{D}_t$.
		\STATE $\{\phi_t, \theta_t\}\leftarrow$ Update the current CVAE using Eq.\eqref{eq:6} with $\mathcal{D}_t$, $p_{\theta_1}$, and $\{y^{:t-1}\}$.
		\STATE Save the decoder, $p_{\theta_t}$, and the labels, $\{y^{:t}\}$.
		\ENDFOR
		\STATE \textbf{return} $p_{\theta_T}\left(x^{:T}|y^{:T}, z\right)$, $\{y^{:T}\}$
	\end{algorithmic}
\end{algorithm}

\section{Experiments}

We evaluated the LGLvKR mainly in two aspects: (1) the error accumulation, and (2) the time complexity on different datasets.

\paragraph{Baseline Methods.}  \label{bm}
There are three types of training strategies in our baseline, fine tuning, joint training, and the existing lifelong training. Specifically, the fine tuning sequentially trains the model with parameters initialized from the recently fine-tuned model on the previous task. The joint training strategy trains the model on the combined real data from tasks seen so far. 
As for the lifelong training, we engaged two widely adopted approaches, pseudo rehearsal and the approach trained only on the dataset of the current task. In particular, both of them were built upon the conditional generative model.
The pseudo rehearsal approaches are implemented on CGAN and CVAE models~\citep{van2020brain,ramapuram2020lifelong,ye2021lifelong}. For CVAE, only the decoder used the label information.
As for the other approaches trained only on the current dataset, we engaged CGAN+EWC \citep{seff2017continual} and CWGAN+RA \citep{wu2018memory}.
In addition, we also compared with the methods that only take the feedback consolidation (LGL-noKR) or knowledge reconstruction (LGL-noFC) on our proposed CVAE where both encoder and decoder use the label information.

\begin{figure}[!th]
	\centering
	\subfigure[]{\includegraphics[width=0.12\linewidth]{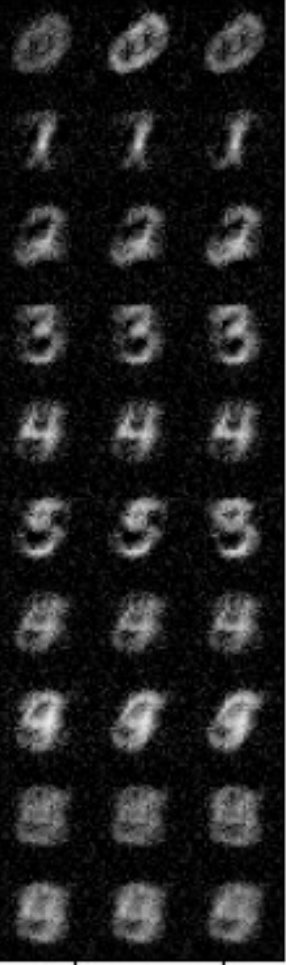} \label{fig:4a} } 
	\subfigure[]{\includegraphics[width=0.12\linewidth]{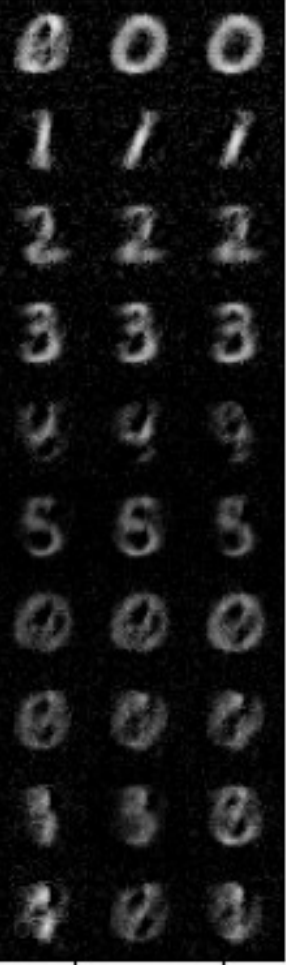} \label{fig:4b}}
	\subfigure[]{\includegraphics[width=0.12\linewidth]{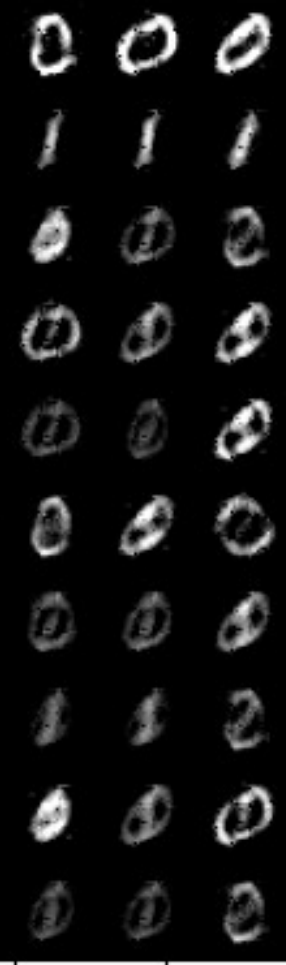} \label{fig:4c}}
	\subfigure[]{\includegraphics[width=0.12\linewidth]{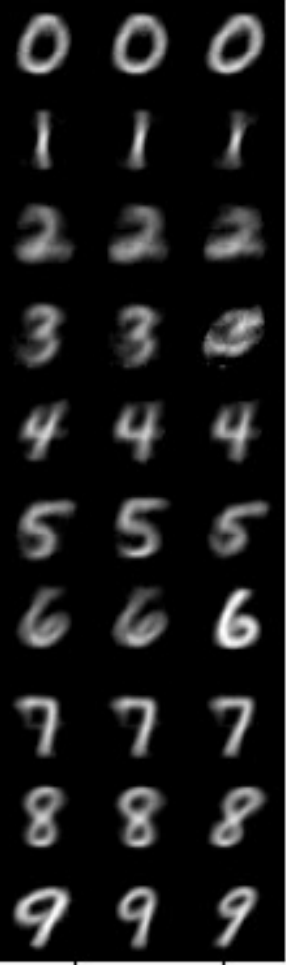} \label{fig:4d}}
	\subfigure[]{\includegraphics[width=0.12\linewidth]{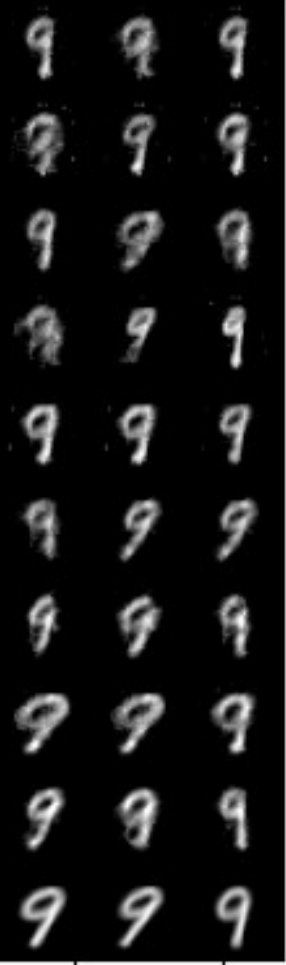} \label{fig:4e}}
	\subfigure[]{\includegraphics[width=0.12\linewidth]{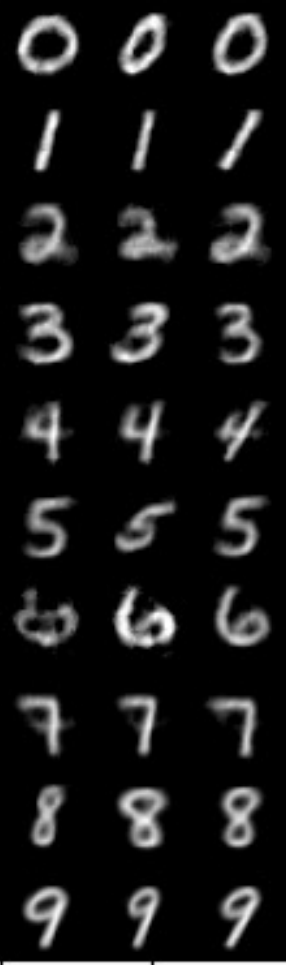} \label{fig:4f}}
	\subfigure[]{\includegraphics[width=0.12\linewidth]{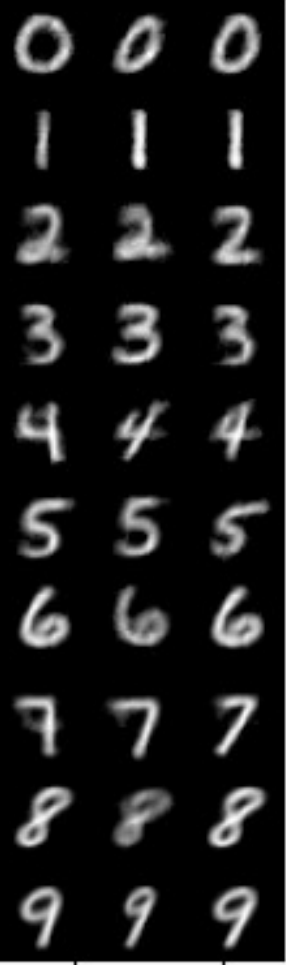} \label{fig:4g}}
	\caption{The lifelong generated digits after sequentially learning $10$ tasks with methods (a) CGAN, (b) CGAN+EWC, (c) CWGAN+RA, (d) CVAE, (e) LGL-noKR, (f) LGL-noFC, and (g) LGLvKR.}
	\label{fig:4}
\end{figure}
\paragraph{Quantitative Metrics.}
We used the accuracy (ACC) and the Fréchet Inception Distance (FID)~\citep{heusel2017gans} as the quantitative metrics.
ACC is the accuracy of the classifier trained on the generated dataset and evaluated on real samples (higher ACC indicates better generation results). FID, instead of evaluating the generated samples directly, compares the statistics between the generated dataset and the real one (lower FID indicates higher generation quality). 

\subsection{Forgetting and Error Accumulation Test} \label{subsec:error}
\paragraph{Experimental Setting.}
To evaluate the effectiveness of our LGLvKR, we consider the lifelong digits generation problem \citep{zhai2019lifelong} in the MNIST \citep{lecun1998mnist} dataset. It consists of $32 \times 32$ pixels images of handwritten digits. Each task contains only one digital category which appears in an ascending order, thus we get $10$ separate tasks. Note that each method takes a small amount of training with only $5$ epochs on each task.
For fairness, we implemented all methods on the network with the same level of parameters (i.e., each model has $0.66$ million parameters). Specifically, CGAN and CGAN+EWC use the same architecture \citep{pytorch2018GAN} while CWGAN+RA takes the original network \citep{liu2020generative}. LGL-noKR, LGL-noFC, and LGLvKR use the networks as shown in Fig.\ref{fig:fig1b} while CVAE has one less layer in encoding label than the network used by LGLvKG. We recorded the average ACC and FID for each method after running $10$ times with different seeds. 
\begin{table*}[!ht]
	\centering
	\begin{tabular}{lc|rl rl rl rl rl rl rl}
		\toprule 
		\multirow{2}*{\makecell{Stra-\\tegy}} &\multirow{2}*{\rotatebox[origin=c]{90}{Task}}	&\multicolumn{2}{c}{CGAN}	&\multicolumn{2}{c}{CGAN+EWC}	&\multicolumn{2}{c}{CWGAN+RA}	&\multicolumn{2}{c}{CVAE}	&\multicolumn{2}{c}{LGL-noKR}	&\multicolumn{2}{c}{LGL-noFC}	&\multicolumn{2}{c}{LGLvKR}	\\
																								\cmidrule(lr){3-4}			\cmidrule(lr){5-6}				\cmidrule(lr){7-8}				\cmidrule(lr){9-10}			\cmidrule(lr){11-12}			\cmidrule(lr){13-14}	\cmidrule(lr){15-16}
																					&			&ACC	&FID				&ACC	&FID					&ACC	&FID					&ACC	&FID				&ACC	&FID					&ACC	&FID		&ACC	&FID			\\
		\midrule
		\multirow{5}{*}{ \rotatebox[origin=c]{90}{Fine Tuning} }	&1		&100	&0.62 	&-	&-	&100	&0.52 	&100	&0.34 	&-	&-	&-	&-	&100	&0.29\\
																					&2		&91.02	&0.67 	&-	&-	&91.16	&0.58 	&93.99	&0.53 	&-	&-	&-	&-	&88.53	&0.55 \\
																					&5		&53.09	&0.84 	&-	&-	&71.58	&0.44 	&62.42	&0.39 	&-	&-	&-	&-	&64.67	&0.38 \\
																				 	&8		&19.00	&0.71 	&-	&-	&54.96	&0.50 	&47.68	&0.44 	&-	&-	&-	&-	&35.80	&0.41 \\
																					&10		&12.90	&0.74 	&-	&-	&48.45	&0.58 	&46.99	&0.43 	&-	&-	&-	&-	&31.99	&0.41 \\
		\midrule
		\multirow{5}*{ \rotatebox[origin=c]{90}{Joint Training} }	&1		&100	&0.69 	&-	&-	&100	&0.53 	&100	&0.39 	&-	&-	&-	&-	&100	&0.37 \\
																	&2		&98.68	&0.79 	&-	&-	&99.80	&0.40 	&99.80	&0.41 	&-	&-	&-	&-	&99.85	&0.41 \\
																	&5		&80.90	&0.59 	&-	&-	&94.45	&0.35 	&95.39	&0.34 	&-	&-	&-	&-	&95.29	&0.34 \\
																 	&8		&58.51	&0.60 	&-	&-	&88.97	&0.35 	&90.34	&0.37 	&-	&-	&-	&-	&90.89	&0.37 \\
																	&10		&51.66	&0.68 	&-	&-	&81.98	&0.32 	&85.85	&0.33 	&-	&-	&-	&-	&86.01	&0.32 \\
		\midrule
		\multirow{10}*{ \rotatebox[origin=c]{90}{Lifelong Training} }	&1		&100	&0.74 	&100	&0.73 	&100	&0.52 	&100	&0.38 	&100	&0.32 	&100	&0.32 	&100	&0.30 \\
																		&2		&99.71	&0.80 	&99.32	&0.73 	&99.76	&0.53 	&\textbf{99.90}	&0.40 	&99.51	&0.51 	&99.76	&0.37 	&99.85	&0.39 \\
																		&3		&94.73	&0.85 	&94.76	&0.66 	&63.09	&0.61 	&96.42	&0.38 	&84.90	&0.42 	&96.88	&0.36 	&\textbf{97.36}	&0.36 \\
																		&4		&91.77	&0.86 	&88.09	&0.73 	&67.77	&0.63 	&93.87	&0.44 	&69.70	&0.55 	&94.53	&0.40 	&\textbf{95.48}	&0.38 \\
																		&5		&82.19	&0.84 	&79.59	&0.55 	&67.68	&0.64 	&93.24	&0.40 	&68.75	&0.42 	&94.75	&0.35 	&\textbf{94.96}	&0.35 \\
																		&6		&79.14	&0.89 	&72.61	&0.71 	&62.25	&0.66 	&90.21	&0.39 	&41.69	&0.35 	&91.42	&0.35 	&\textbf{91.44}	&0.34 \\
																		&7		&68.29	&0.93 	&65.05	&0.67 	&63.08	&0.67 	&89.61	&0.41 	&52.04	&0.42 	&90.12	&0.35 	&\textbf{90.54}	&0.32 \\
																		&8		&55.24	&0.95 	&60.36	&0.70 	&49.16	&0.68 	&88.90	&0.46 	&50.71	&0.48 	&89.82	&0.36 	&\textbf{89.84}	&0.35 \\
																		&9		&53.06	&0.93 	&34.99	&0.77 	&46.82	&0.68 	&85.49	&0.49 	&37.83	&0.57 	&86.18	&0.36 	&\textbf{86.58}	&0.33 \\
																		&10		&46.44	&0.90 	&32.65	&0.71 	&44.20	&0.69 	&82.67	&0.51 	&43.06	&0.47 	&82.27	&0.33 	&\textbf{83.94}	&0.31 \\
		\bottomrule
	\end{tabular}
	\caption{The ACC (\%) and FID values of different methods with three training strategies.}
	\label{tab:error}
\end{table*}

\paragraph{Analysis.}

We analyze the results from three aspects. Firstly, in Tab.\ref{tab:error} and~Fig.\ref{fig:4a}, fine tuning the generative networks (used in CGAN, CWGAN+EWC, CVAE, LGLvKR) will incur the catastrophic forgetting problem since its ACC dramatically degrades from $100\%$ to the level below $50\%$. Correspondingly, the FID value substantially increases, which means the quality of the generated images decreases.

Secondly, we see that the joint training gets good results, which are the upper bounds of lifelong learning. It confirms that the decline of fine tuning results is not limited by the network capacity but by the catastrophic forgetting problem. Besides, by comparing the results of VAE-based models (CVAE, LGLvKR) and GAN-based models (CGAN, CWGAN+RA), we found that the methods based on VAE relatively outperform methods based on GAN. That is mainly because the GAN-based models are not well trained with $5$ epochs while it is enough for the VAE-based model, indicating the high efficiency in the training process of VAE. On the other hand, by comparing CVAE and LGLvKR, a relative improvement could be obtained by embedding $y$ in both encoder and decoder. It is prominent when there are more categories, such as $86.01$ for LGLvKR versus $85.85$ for CVAE after learning $10$ tasks.

Finally, for the lifelong training, LGLvKR achieved the best ACC, almost reaching its upper bound. In contrast, the CVAE method using pseudo rehearsal also obtained considerable ACC, but its FID results are not good, which was also verified by the generated images. As shown in Fig.\ref{fig:4d} and Fig.\ref{fig:4g}, the generated digits (from $1$ to $5$) of CVAE are more blur compared with those of LGLvKR. The GAN-based methods performed worse. First, they could not be well trained in $5$ epochs. Besides, error accumulation heavily influences their results. Especially for the CWGAN+RA, its ACC dramatically degraded to $44.20$ while its model reached $81.98$ with joint training. In Tab.\ref{tab:error}, Fig.\ref{fig:4a}, and Fig.\ref{fig:4b}, CGAN and CGAN+EWC obtained appropriate performance against a few task sequences but failed when the task length exceeds $5$.

We also conducted ablation studies. As shown in Tab.\ref{tab:error} and Fig.\ref{fig:4e}, LGL-noKR showed the similar catastrophic forgetting problem since it trained with CVAE loss of Eq.\eqref{eq:3} and the feedback consolidation loss of Eq.\eqref{eq:5}. By expending the intrinsic reconstruction character of VAE to the knowledge reconstruction, LGL-FC well retained the historical tasks seen so far. And it was further improved with the help of feedback consolidation, which came to the LGLvKR.

\begin{figure}[!th]
	\centering
	\includegraphics[width=0.49\textwidth]{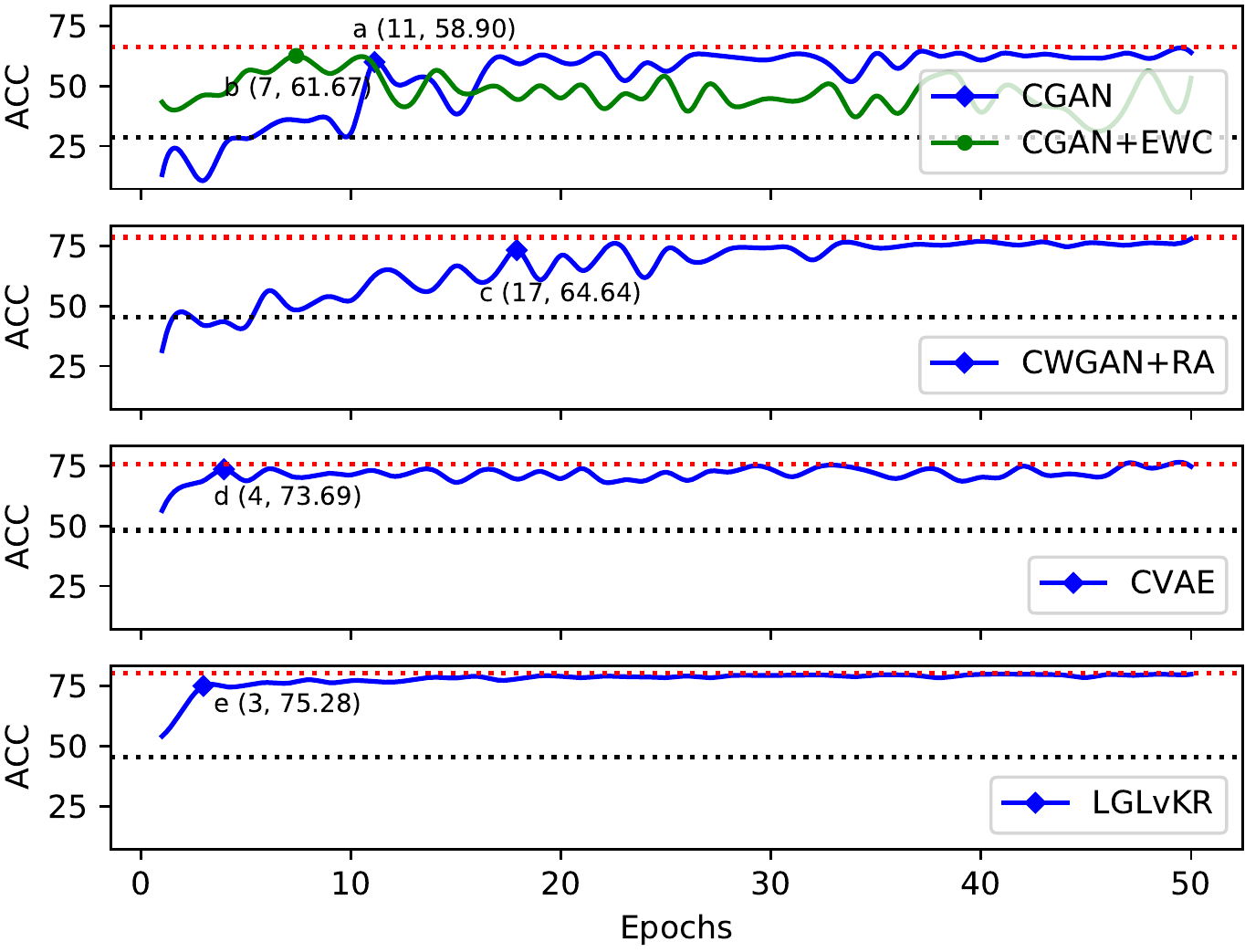}
	\caption{ACC of various lifelong generative methods after sequentially training $10$ separate fashion-MNIST tasks. The black and red dash lines indicate, respectively, their corresponding lower and upper bounds.}
	\label{fig:fig5}
\end{figure}

\subsection{Time Consumption Test}
\begin{figure}[!th] 
	\centering
	\subfigure[]{\includegraphics[width=0.15\linewidth]{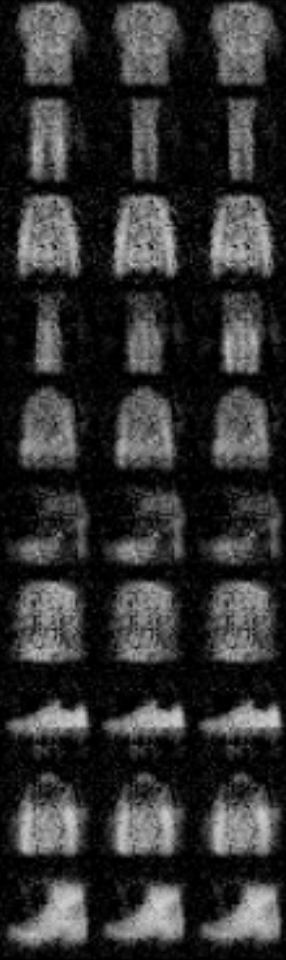} \label{fig:6a}}
	\subfigure[]{\includegraphics[width=0.15\linewidth]{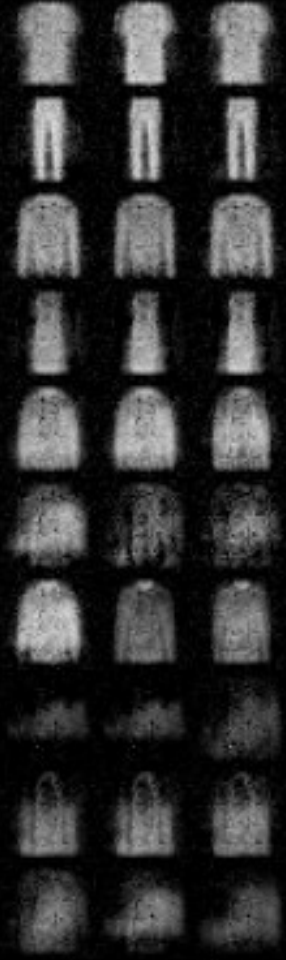} \label{fig:6b}}
	\subfigure[]{\includegraphics[width=0.15\linewidth]{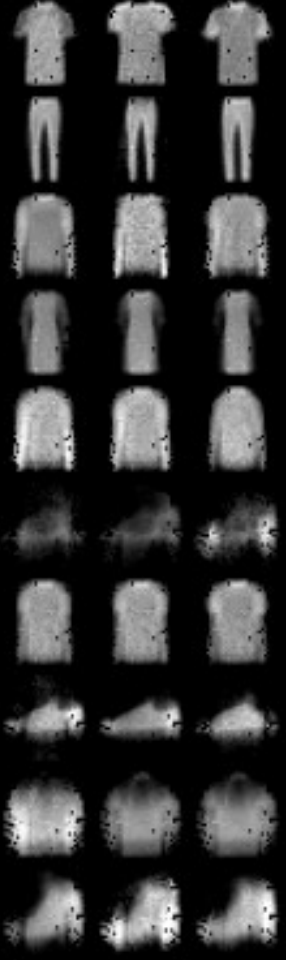} \label{fig:6c}}
	\subfigure[]{\includegraphics[width=0.15\linewidth]{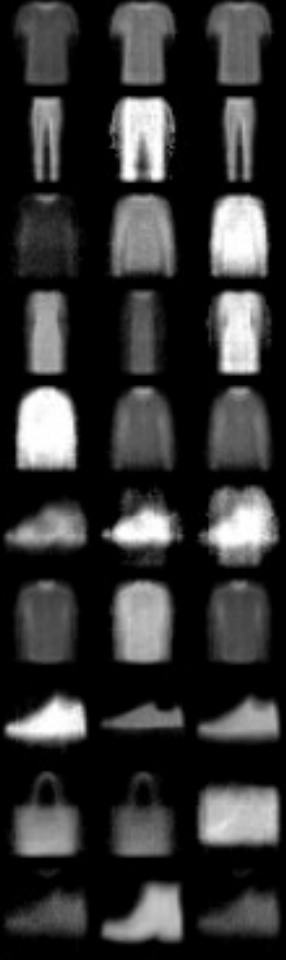} \label{fig:6d}}
	\subfigure[]{\includegraphics[width=0.15\linewidth]{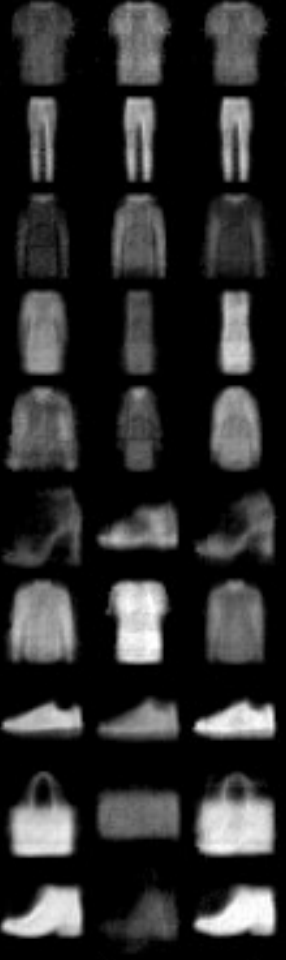} \label{fig:6e}}
	\caption{The generated images from methods (a) CGAN ($700$s), (b) CGAN+EWC ($388$s), (c) CWGAN+RA ($1230$s), (d) CVAE ($217$s), and (e) LGLvKR ($157$s), with their average training time (seconds). We took the models to generate images when the methods first reached the best ACC, while the best ACC was stressed in Fig.\ref{fig:fig5}.}
	\label{fig:6}
\end{figure}
\paragraph{Experimental Setting.}
We evaluate the efficiency of LGLvKR on the fashion-MNIST dataset~\citep{xiao2017fashion}. 
Specifically, we tested the ACC to the number of epochs (from $1$ to $50$) trained on each task and the time required to achieve comparable results.
Similar to Section \ref{subsec:error}, we implemented each method on the network with the same level of parameters. And the fashion-MNIST dataset was divided into $10$ separate tasks. 
We took the best results of fine tuning and joint training over $50$ epochs as the lower and upper bounds of ACC, respectively. 
With the fixed epoch $i$, each model trains $i$ epochs on each task after sequentially training $10$-tasks. We saved the trained generative model. Then, a classifier was trained on the samples generated from the saved model until convergence. We finally took the classifier to calculate the ACC on the real test dataset. Each experiment runs $10$ times with different seeds. 

\paragraph{Analysis.}
Since CGAN and CGAN+EWC engaged the same network, they shared the same upper and lower bounds. As shown in Fig.\ref{fig:fig5}, CGAN and CGAN+EWC first reached their optimal ACC, with $11$ and $7$ epochs training, respectively.
Due to the nature drawback of unstable training \citep{arjovsky2017wasserstein}, CGAN was unstable as the number of epochs grows. In particular, this unstableness pronounced for CGAN+EWC owing to its poor performance on the long length of sequent tasks.
By introducing the Wasserstein metric and emphasizing the importance of label in the discriminator \citep{wu2018memory,liu2020generative}, CWGAN+RA obtained a higher upper bound of ACC (66.45\% for CGAN/CGAN+EWC versus 78.66\% for CWGAN+RA) and more stable results on lifelong generating learning. However, CWGAN+RA conducted $17$ epochs to reach the best ACC for the first time.
Compared with CWGAN+RA, although CVAE got an inferior upper bound in joint training (75.88\% for CVAE versus 78.66\% for CWGAN+RA), it obtained more stable results benefiting from its stable training mechanism. CAVE took $4$ epochs to reach the optimal ACC for the first time. 
By embedding the label information in both the encoder and decoder, LGLvKR achieved a higher ACC upper bound (80.47\%) compared with CVAE (75.88\%) with the joint training strategy. On the other hand, instead of pseudo rehearsals, LGLvKR took knowledge reconstruction to retain the historical knowledge and was enhanced by a feedback consolidation. It got both the optimal ACC with only $3$ epochs and a considerable improvement in stability.
As demonstrated in Fig.\ref{fig:6},
overall, LGLvKR achieved lifelong learning with the shortest time while getting a good picture of diversity. 
CVAE in Fig.\ref{fig:6e} whereas confused Sandal (the sixth class) and Ankle boot (the last class).

\begin{figure}[!t] 
	\centering
	\subfigure[]{\includegraphics[width=0.19\linewidth]{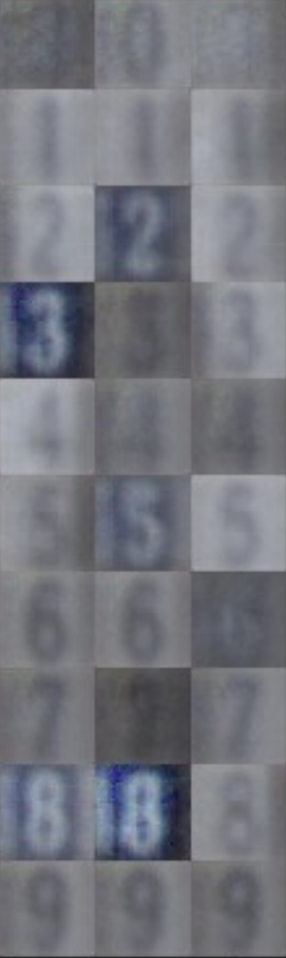} \label{fig:7a}}
	\subfigure[]{\includegraphics[width=0.19\linewidth]{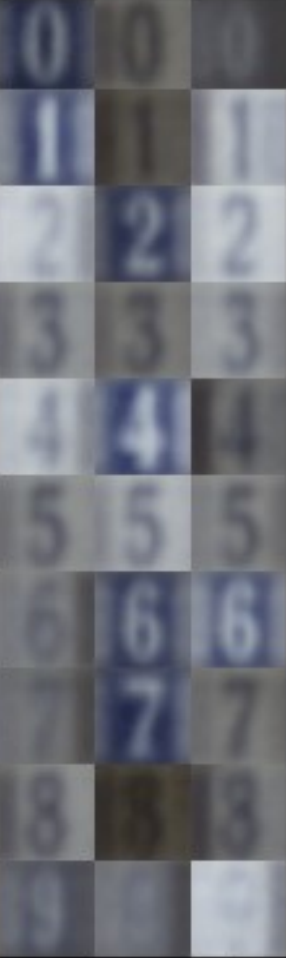} \label{fig:7b}}
	\subfigure[]{\includegraphics[width=0.19\linewidth]{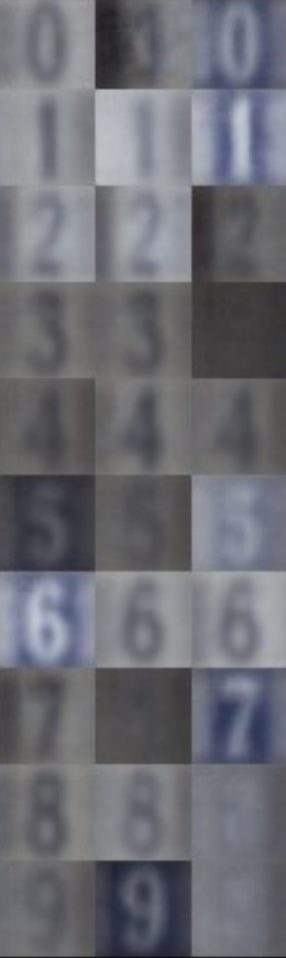} \label{fig:7c}}
	\subfigure[]{\includegraphics[width=0.19\linewidth]{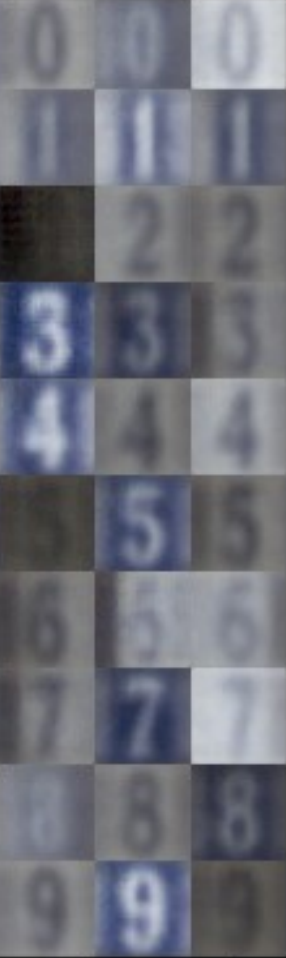} \label{fig:7d}}
	\caption{The generated colored images from CVAE with lifelong training (a) and joint training strategy (b), along with LGLvKR with lifelong training (c) and joint training (d).}
	\label{fig:7}
\end{figure}

Finally, we showed the generated images by CVAE and LGLvKR on the colored Street View House Number~\citep{netzer2011reading}. In particular, we trained each model with lifelong training and joint training strategies. 
We referred \citep{pytorchVAE} to implement the models.
Other experimental settings are similar to those in fashion-MNIST while we fixed the training epochs with $10$. As shown in Fig.\ref{fig:7}, the FID values of all the images were $0.40$ (Fig.\ref{fig:7a}), $0.36$ (Figs.\ref{fig:7b}), $0.35$ (Fig.\ref{fig:7c}), and $0.32$(Fig.\ref{fig:7d}), respectively. The training time of the corresponding methods are, respectively, $1185$, $4870$, $1059$, and $4885$ seconds. 
Overall, LGLvGR achieved considerable results with the shortest training time.


\section{Conclusion}

We studied the lifelong generative learning problem based on a variant of the CVAE. By expending the intrinsic reconstruction character of VAE to reconstruct the historical knowledge learned before, LGLvKR could well handle the catastrophic forgetting problem. In addition, to alleviate the error accumulation, we further developed a feedback strategy for LGLvKR. Experiments on MNIST, FashionMnist, and SVHN verified that the proposed LGLvKR addressed the lifelong generating problem effectively and efficiently.

This work mainly focuses on verifying that the intrinsic reconstruction character of VAE is good enough to retain the historical knowledge and we extend it to lifelong generating learning. However, the lifelong generating output of a more complex image is still an open challenge~\citep{lesort2020continual}. Thus, in the future, it is desirable to introduce a more generalized architecture of the network in the LGLvKR framework. Also, one alternative direction is to generate the high-level features from the complex images in lifelong learning.



\bibliographystyle{named.bst}
\bibliography{ijcai22.bib}

\end{document}